\documentclass{egpubl}
\usepackage{pg2023}

\SpecialIssuePaper         

\CGFStandardLicense

\usepackage[T1]{fontenc}
\usepackage{dfadobe}  

\usepackage{cite}  
\BibtexOrBiblatex
\electronicVersion
\PrintedOrElectronic
\ifpdf \usepackage[pdftex]{graphicx} \pdfcompresslevel=9
\else \usepackage[dvips]{graphicx} \fi

\usepackage{egweblnk} 

\usepackage{amsfonts}
\usepackage{times}
\usepackage{graphicx}
\usepackage{mathtools}
\usepackage{booktabs}
\usepackage{makecell}
\usepackage{pdfpages}

\usepackage{algorithm}
\usepackage{algorithmic}

\usepackage{caption,subcaption}
\DeclarePairedDelimiter\floor{\lfloor}{\rfloor}
\DeclareMathOperator*{\argmin}{arg\,min}

\title[Neural Representation of Volumetric Data]%
      {Efficient Neural Representation of Volumetric Data using Coordinate-Based Networks. }

\author[S. Devkota \& S. Pattanaik]
{\parbox{\textwidth}{\centering S. Devkota and S. Pattanaik
        }
        \\
{\parbox{\textwidth}{\centering University of Central Florida, Orlando FL, USA
       }
}
}

\begin{document}

\teaser{
 \includegraphics[width=\linewidth]{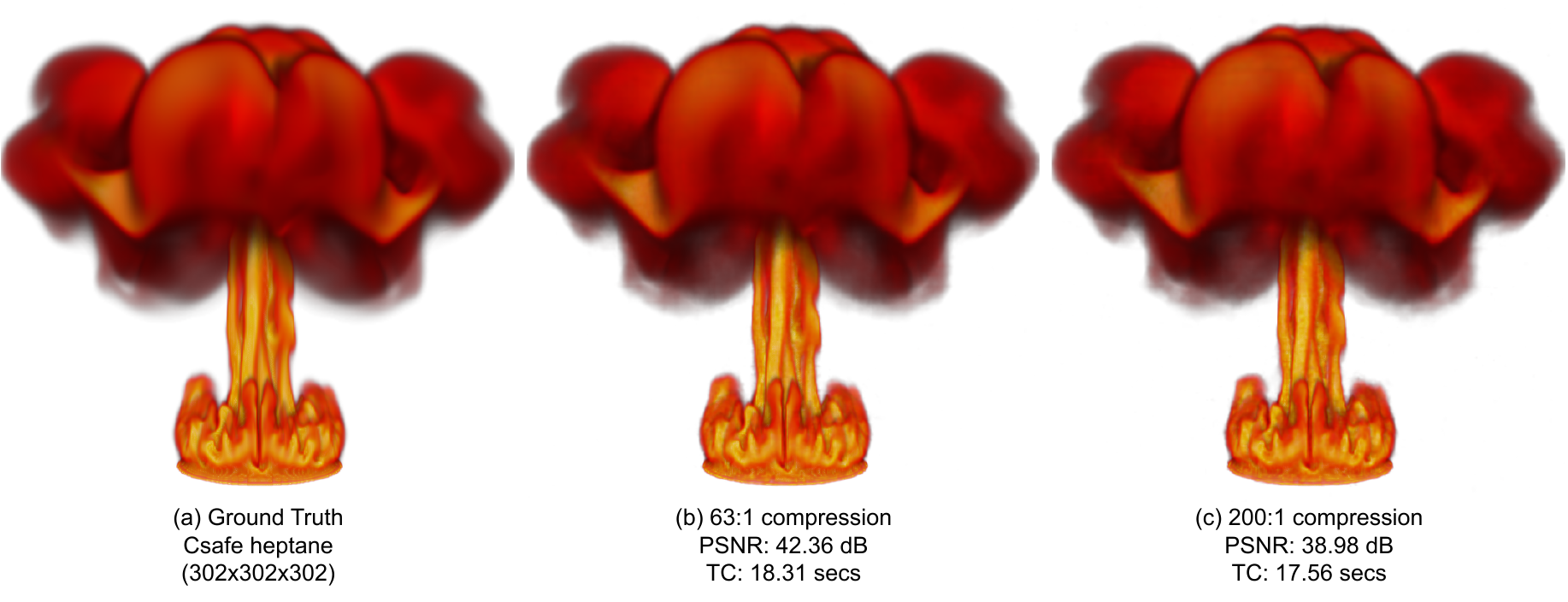}
 \centering
  \caption{Neural representation of volumetric data using our proposed method. (PSNR: Peak Signal to Noise Ratio, TC: Time to compress)}
\label{fig:teaser}
}

\maketitle
\begin{abstract}
In this paper, we propose an efficient approach for the compression and representation of volumetric data utilizing coordinate-based networks and multi-resolution hash encoding. Efficient compression of volumetric data is crucial for various applications, such as medical imaging and scientific simulations. Our approach enables effective compression by learning a mapping between spatial coordinates and intensity values. We compare different encoding schemes and demonstrate the superiority of multi-resolution hash encoding in terms of compression quality and training efficiency. Furthermore, we leverage optimization-based meta-learning, specifically using the Reptile algorithm, to learn weight initialization for neural representations tailored to volumetric data, enabling faster convergence during optimization. Additionally, we compare our approach with state-of-the-art methods to showcase improved image quality and compression ratios. These findings highlight the potential of coordinate-based networks and multi-resolution hash encoding for an efficient and accurate representation of volumetric data, paving the way for advancements in large-scale data visualization and other applications.


\end{abstract}  

\section{Introduction}
\label{intro}

Visualization of large-scale data can be a challenging problem. First, large-scale data sets can require significant amounts of storage space, making it difficult to store and access the data efficiently. Second, large-scale data sets can take a long time to transfer over a network, which can impact the performance of visualization. Another challenge with large-scale data visualization is that the sheer size of the data can overwhelm visualization tools and systems, leading to slow rendering times, unresponsive interfaces, and difficulty in exploring the data.

Data compression can help address some of these challenges by reducing the amount of data that needs to be stored, transferred, and processed. There are many different techniques that can be used for volume data compression, ranging from traditional methods such as lossless and lossy compression to more recent approaches based on deep learning and neural networks. 

One of the earliest methods for lossy compression was introduced in the 1990s by Ning and Hesselink \cite{Ning92}, who proposed vector quantization (VQ) as a way to compress 3D scalar data. In their method, a codebook of representative 3D vectors is constructed by clustering similar data points. The original data is then compressed by replacing each data point with the closest codebook vector. Other early research on lossy compression for volume rendering also includes using discrete cosine transform (DCT) based compression \cite{Chan91} \cite{Yeo95} and Fourier domain-based volume rendering \cite{Malzbender93} \cite{Levoy92} \cite{Totsuka93}. These techniques are particularly useful for compressing smooth data, where most of the energy is concentrated in low-frequency components. Since the human visual system is more perceptive to low-frequency components, the high-frequency components from the image are safely discarded while still maintaining most of the information contained in the image.

Recently, scene representation using neural networks have gained a lot of popularity in the visualization community where the network encodes a field function that maps input 3D coordinates (coupled with direction vectors in some cases) to a field value, such as density or radiance, using neural networks. This approach has enabled a range of applications, such as novel view reconstruction \cite{Mildenhall20} and compression \cite{Strumpler21} \cite{Takikawa22}. Thanks to the flexibility and differentiability of neural networks, this new approach has opened up many possibilities for volume data compression and visualization \cite{neurcomp} \cite{Weiss2}.

In our work, inspired by scene representation networks (SCN), we plan to represent volume data by directly approximating the mapping from spatial coordinates to volume values using a multi-layer perceptron (MLP). The trained MLP is then considered a compressed version of the original data. This representation is efficient because the memory footprint of a neural network is often orders of magnitude smaller than the original data, and sampling the representation is flexible as one can arbitrarily query volume values without explicit decompression and interpolation.

Recent research has demonstrated the efficacy of optimization-based meta-learning in reducing the number of gradient descent steps required for optimizing coordinate based neural networks in various domains, including images\cite{Strumpler21}, signed distance fields\cite{metasdf}, and radiance fields\cite{Tancik20b}. In our work, we focus on learning the weight initialization for neural representations specifically tailored to volumetric data. By utilizing learned values as the initial network weights, we establish a strong prior that facilitates faster convergence during optimization. To achieve this, we employ Reptile \cite{reptile}, an optimization-based meta-learning algorithm, to generate optimized initial weights for representing a specific signal class, such as medical volume datasets. We opt for Reptile over alternative meta-learning algorithms like MAML \cite{MAML} due to its simplicity and computational efficiency. While MAML differentiates through the computation graph of the gradient descent algorithm, Reptile employs gradient descent individually on each task, eliminating the need for graph unrolling or second derivative calculations \cite{reptile}. This key distinction allows for lower memory consumption and better computational efficiency.

To summarize, the main contributions from our work are as follows:
\begin{itemize}
     \item A fast neural compression approach for volume data based on multiresolution hash encoding and a study on performance comparison with other input encoding techniques. 
     \item Evaluation on the effectiveness of representation transfer with meta-learned intialization for neural compression of volume data. 
     \item Experimental results on the efficacy of our volume compression approach against other state of the art techniques
\end{itemize}

\begin{figure*}[ht]
  \centering
  \includegraphics[width=\linewidth]{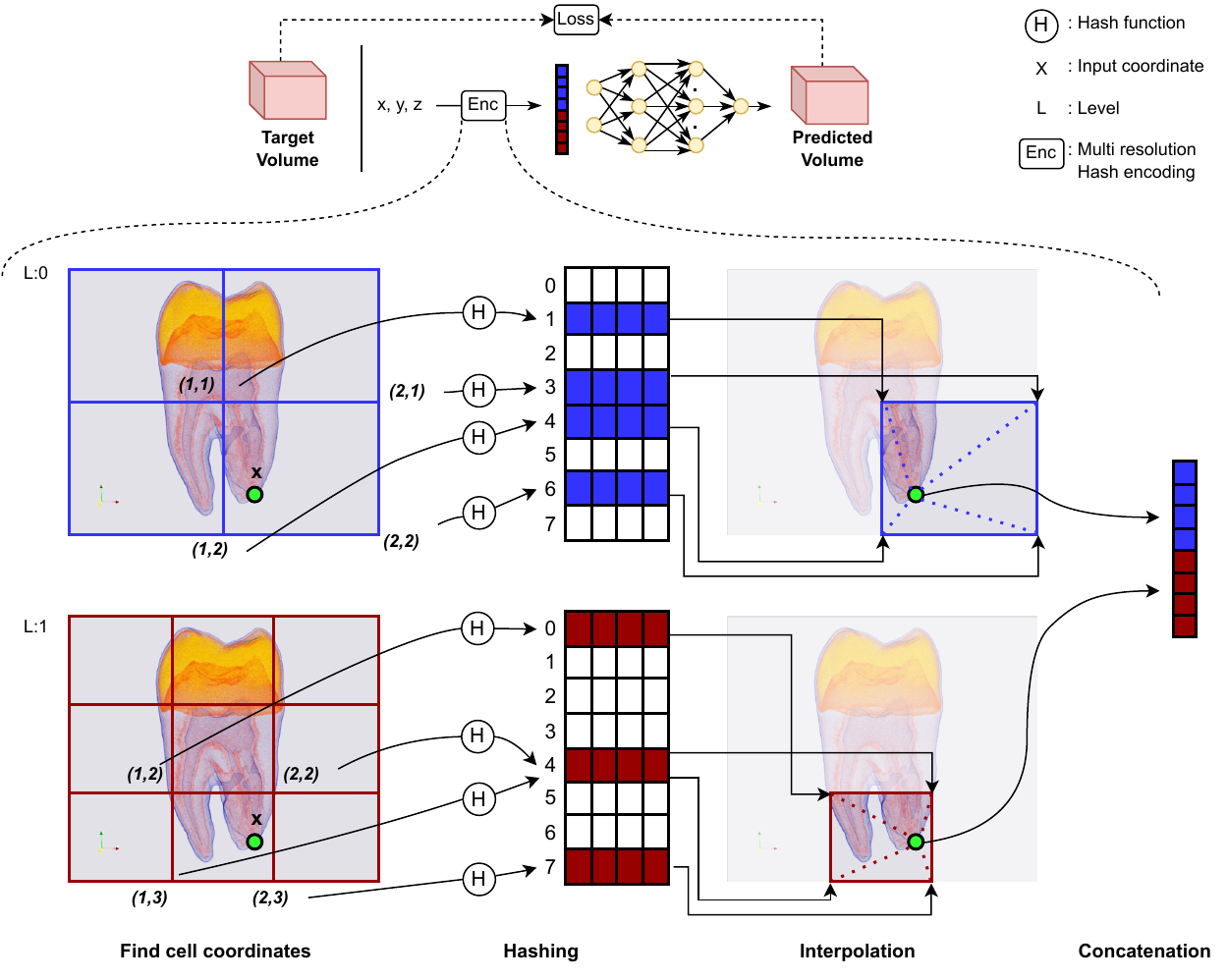}
  \centering
  \caption{Overview of Multi-resolution hash encoding. In this two-dimensional example, we first break the image in $L$ resolution levels (grids). Figure shows an example with two levels, $L:0$ and $L:1$. For a given normalized input coordinate $x$, the integer coordinates of the surrounding corners are hashed to obtain an index to a hash table with size $T$. Every entry in the hash table is a trainable feature vector of size $W$. In the example shown above, we have $L=2$, $W=4$, and $T=8$. The feature vectors from the surrounding corners are linearly interpolated to obtain the feature vector at coordinate $x$. Then all the feature vectors for x from each level are concatenated with each other which forms the final encoded vector for input $x$.}   
  \label{fig:hashencoding}

\end{figure*}

\section{Related Work}
\label{rw:neural-volume}

Volume data compression has been extensively researched in the field of computer science and data compression. Various lossy and lossless techniques have been proposed to achieve high compression ratios while preserving the quality of data. In this section, we provide an overview of some of the key works in three different research areas that are closely tied with each other: volume compression, neural scene representation, and neural volume representation.

\subsection{Volume Compression}
\label{rw:volumecompression}
Recent years have seen a growing interest in lossy compression of large-scale volumes, driven by the expanding capabilities of simulation and data acquisition. Early research focused primarily on discrete cosine transforms \cite{Chan91} \cite{Yeo95}, and wavelet-based compression \cite{Park99} \cite{Muraki93}. These techniques involve removing high-frequency information to achieve sparser data representation, which can then be quantized to reduce the number of bits required to store the data. Quantization-based volume compression techniques \cite{Ning92} \cite{Gelder96} \cite{Levoy96} \cite{Tarini00} involve dividing the volume into non-overlapping regions and mapping each region to a discrete value a set of representative values, which can be either pre-defined or learned from the data \cite{Schneider03} \cite{Fout07} \cite{Guthe16}.

Inspired by fixed-rate texture compression methods, ZFP \cite{ZFP} provides a compression method for floating point data in multi-dimensional arrays. This method uses a fixed-rate compression approach that maps small blocks of values in multiple dimensions to a user-specified number of bits per block. In recent years, tensor decomposition-based compression techniques \cite{Ballester15}\cite{TAMRESH} are also being used for compactly encoding large multidimensional arrays. A notable mention in this category is TTHRESH \cite{TTHRESH} which utilizes Higher Order Singular Value Decomposition (HOSVD) with bit-plane, run-length, and arithmetic coding for volume data compression.

\subsection{Scene Representation using Neural Networks}
\label{rw:scenerepresentation}
Scene representation networks (SRN) are a class of deep neural networks that can be used to encode occupancy fields \cite{Mescheder19}\cite{Peng20}, implicit surfaces like SDF \cite{Martel21}\cite{Michalkiewicz19}\cite{Park19}, or radiance fields \cite{Liu20}\cite{Mildenhall20}\cite{Muller21} of complex 3D scenes in a compact and efficient manner. Most of these approaches use an encoding method that maps the input coordinates to a higher dimensional space before passing them to the network. For instance, to synthesize novel views of complex scenes using a sparse set of input images, Mildenhall \emph{et al.} proposed to encode coordinates as a multi-resolution sequence of sine and cosine functions for the NeRF algorithm \cite{Mildenhall20}. Later, it was shown that using sinusoids with logarithmically-spaced axis-aligned frequencies further improves the reconstruction ability of coordinate-based networks \cite{Tancik20}. These encoding schemes are referred to as frequency-based encoding.  

Recently, parametric encodings have been introduced which achieve state-of-the-art results. This encoding scheme involves arranging additional trainable parameters in auxiliary task-specific data structures such as a tree \cite{Takikawa21} or grid \cite{Jiang20} \cite{Sun21} \cite{Yu21}. 
Although the total number of trainable parameters is higher for parametric encoding, it enables the use of much smaller coordinate-based networks and can be trained to converge much faster without sacrificing approximation quality. Recently, M{\"{u}}ller \emph{et al.} proposed the use of a multiresolution hash table of trainable parameters for input encoding which is task-independent and outperforms all the previous approaches \cite{instantnerf}. Motivated by this, we plan to implement this encoding scheme in our volume representation.

While our approach draws inspiration from the principles of M{\"{u}}ller \emph{et al.}'s work \cite{instantnerf}, there are notable differences that distinguish our approach. Unlike their work, we focus on neural representation of volumetric data where the network maps input coordinates to voxel values. Additionally, our study incorporates meta-learning to facilitate more efficient parameter initialization for the neural network representation. This inclusion of meta-learning allows us to harness domain-specific knowledge and further optimize the compression process for volumetric data.

\subsection{Volume Representation using Neural Networks}
\label{rw:volumerepresentation}
Several approaches have been proposed to use deep learning in representing volume data for visualization. Some of the previous works introduced a new technique for volume visualization, which does not rely on the traditional rendering pipeline. Instead, they used either Generative Adversarial Networks \cite{Berger19} or encoder-decoder \cite{Hong19}. He \emph{et al.} \cite{He19} used a convolutional regression model to learn the mapping from the simulation and visualization parameters to the final visualization. This allowed flexible exploration of parameter space for large-scale ensemble simulations. These networks can render the volume data directly using the compact representations stored in the network weights. However, the networks may not perform well if the training data does not contain the specific combination of views and transfer functions that are used in the test data.

The field of super-resolution is closely related to volume compression and can be used to enhance the visual quality of low-resolution volume data or rendered frames. Instead of storing data in higher resolution, super-resolution networks can efficiently upscale the resolution of the data. One approach by Weiss \emph{et al.} \cite{Weiss19} used a deep learning-based architecture for isosurface rendering. Another method by Devkota \emph{et al.} \cite{Devkota22} implemented temporal reprojection for volumetric cases to perform super-resolution for direct volume rendering. Although these methods were applied to volumetric scalar fields, other works have focused on temporal \cite{TSR-TVD} and spatial \cite{SSR-TVD} super-resolution for time-varying vector field data. Additionally, recent advancements in neural representation have led to methods that handle diverse scientific visualization tasks in a single framework. For instance, Han \emph{et al.} \cite{CoordNet} proposed a unified coordinate-based neural network architecture capable of tackling both super-resolution and visualization tasks relevant to time-varying volumetric data visualization.

Recent contributions have brought forward numerous SCN-based approaches for volume representation. Neurcomp \cite{neurcomp} demonstrated the effectiveness of coordinate-based networks for volume compression for scalar field data. Using implicit neural representation, their approach frames compression as function approximation, achieving highly compact representations that outperform existing volume compression methods. Their utilization of neural networks introduces a novel way of handling scalar field compression, and its performance benchmarks will be used for direct comparisons in our evaluation.
Weiss \emph{et al.}\cite{Weiss2} proposed to use SCNs with tensor cores for faster decoding time and lower memory consumption during data reconstruction. Kim \emph{et al.} introduced NeuralVDB \cite{neuralvdb}, a hybrid storage scheme with hierarchical neural network and wide VDB tree structure for efficient storage of sparse volumetric data. These studies have shown that SCN-based compression method achieves high compression ratios while maintaining a high level of visual fidelity compared to traditional compression methods. This makes it a promising approach for compressing large volumetric datasets in scientific visualization and medical imaging applications.

\section{Methodology}
\label{nvr:methodology}

We are interested in a function $\phi_\theta(x)$ that satisfies the objective function of the form
\begin{equation}\label{eq:objective}
\argmin_\theta F (\delta_x , \phi_\theta(x))
\end{equation}

The function $\phi_\theta(x)$ is a coordinate-based network with parameters $\theta$ that maps the spatial coordinates $x \in \mathbb{R}^d$ to some value that is as close as possible to $\delta_x$, which is the intensity at coordinate $x$. Thus, our goal is to define such a volume representation network $\phi_\theta(x)$ that takes spatial coordinates as input to learn a mapping between the input coordinates and a target output.

In coordinate-based scene representation networks, the input coordinates undergo an encoding process to transform them to a higher-dimensional space before being inputted into the neural network. The encoding of the input plays a significant role in capturing the spatial information of the input data. If the input is not encoded, the network can only learn a relatively smooth function of position, which results in an inadequate representation of the intensity field \cite{instantnerf}.

Recently, state-of-the-art results have been achieved by parametric encodings \cite{Jiang20}\cite{Sun21}\cite{Yu21} for coordinate based networks. Motivated by these works, we propose to use multi-resolution hash encoding \cite{instantnerf} which we explain in the following section.


\subsection{Multi-resolution Hash Encoding}
\label{section:multi-res-hash}

Multi-resolution Hash Encoding introduces a multi-step process to effectively encode input coordinates. This subsection details each step of the encoding process.

Step 1 - Find cell coordinates

The encoding starts with breaking up a $d$ dimensional array into $L$ different levels (grids) with increasing resolution. Figure \ref{fig:hashencoding} shows an example domain with two dimensions (an image). The resolution for each level is a constant multiple of the previous level. The constant multiple is given by a growth factor $b$, which is computed based on three hyper-parameters: number of levels $L$, resolution of the coarsest level $N_0$ and resolution of the finest level $N_{L-1}$

\begin{equation}\label{eq:grownfactor}
b = e^\frac{\ln{N_{L-1}}  - \ln{N_0}}{L-1}
\end{equation}

Therefore, the resolution for each grid level $l$ is computed as 
\begin{equation}\label{eq:resolution}
N_l = \floor*{ N_{l-1} * b} = \floor*{N_0 * b^l}.
\end{equation}

Consider a normalized input coordinate $x \in \mathbb{R}^d$ in [0, 1]. For each level $l$, $x$ is first scaled by the resolution of that level and is rounded up and down to find the integer coordinates of each corner of the cell containing $x$.

Step 2 - Hashing

Each of the $2^d$ integer coordinates surrounding the coordinate $x$ is hashed using a hash function $H(x)$ \cite{hashfunc}.
\begin{equation}\label{eq:hash}
H(x) = \bigoplus_{i=1}^{d} x_i \pi_i  \bmod  T
\end{equation}
Here, $\pi_i$ denotes large prime numbers for each dimension $i$ and $\bigoplus$ denotes bit-wise XOR operation. 
The output from the hashing function is the index to a trainable hash table of size $T$. Each entry in the hash table is an array of trainable weights of size $W$. Thus, each grid level $l$ has a corresponding hash table which is described by two hyper-parameters $W$ and $T$. During training, the gradients are back-propagated all the way back to the hash table entries, dynamically optimizing them to learn a good input encoding.

Step 3 - Interpolation

After all the surrounding coordinates are mapped to index values which is used to query the hash table, the trainable weights from the corresponding slots are linearly interpolated in $d$ dimensions to obtain the feature vector at coordinate $x$.

Step 4 - Concatenation

These feature vectors of size $W$ from each of the $L$ levels are concatenated together to form the input to a multi-layer perceptron. The output from the MLP is the predicted intensity value at coordinate $x$.



\subsection{Meta Learned initialization}
When we overfit any model to a volume data sample, it learns their parameters from scratch during training. With weights typically initiated at random, these models do not carry any domain knowledge about the volume data they are assigned to compress. Simply put, these models suffer from a lack of inductive biases. Inductive biases are assumptions that a model makes about the data that can help the model to learn more effectively.

To address this, we propose to apply the Reptile algorithm \cite{reptile}. This first-order meta-learning algorithm works by consistently selecting a task, conducting stochastic gradient descent on the chosen task, and subsequently shifting the initial parameters in the direction of the final parameters that were learned during that task. By building on inductive biases, it aims to enrich the model with a greater understanding of the data domain before the actual learning process begins.

In our case, we start with multiple volume datasets, sourced from a specific distribution $V_d$ (for example: medical imaging or scientific simulation). The goal here is to identify initial weights, represented by $\theta$, which would yield the lowest possible loss when we optimize a network to represent a novel and previously unseen volume from the same distribution.

For our purpose, the algorithm works as shown in \ref{alg:reptile}

\begin{algorithm}[ht]
\caption{Reptile Meta-Learning \cite{reptile}}
\label{alg:reptile}
\begin{algorithmic}[1]
\STATE \textbf{Initialize} $\theta$, the initial parameter vector
\FOR{iteration $1,2,3,\ldots$}
\STATE Select a volume $V$ at random from distribution $V_d$
\STATE Perform $k > 1$ steps of gradient descent on $V$, starting with parameters $\theta$, resulting in parameters $W$
\STATE \textbf{Update:} $\theta \gets \theta + \epsilon(W - \theta)$
\ENDFOR
\RETURN $\theta$
\end{algorithmic}
\end{algorithm}

\section{Experiments}

In this section, we perform an array of experiments, ranging from investigating different encoding schemes and hyperparameters to performance comparisons against state-of-the-art techniques. In all of our experiments, we adopt mixed precision training \cite{mixedprecision}, where the neural network weights are stored as $float16$. During training, in order to match the accuracy of the $float32$ networks, a $float32$ master copy of weights is maintained for parameter update. Unless otherwise stated, our hash encoding network in all of our experiments employs an MLP with two hidden layers and 64 neurons. Additionally, we use ReLU (rectified linear unit) as the activation function and $L2$-loss to guide the training process.

\subsection{Encoding Schemes}

\begin{figure*}[ht]
    \centering
    \begin{subfigure}{0.49\textwidth}
        \centering
        \includegraphics[width=\textwidth]{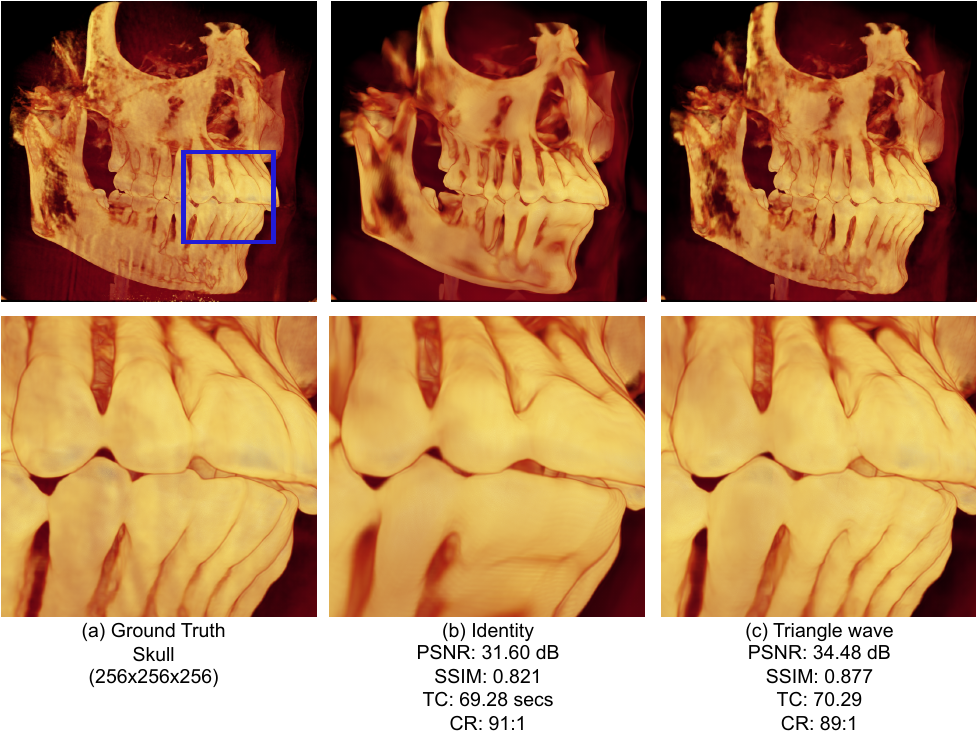}
    \end{subfigure}
    \begin{subfigure}{0.49\textwidth}
        \centering
        \includegraphics[width=\textwidth]{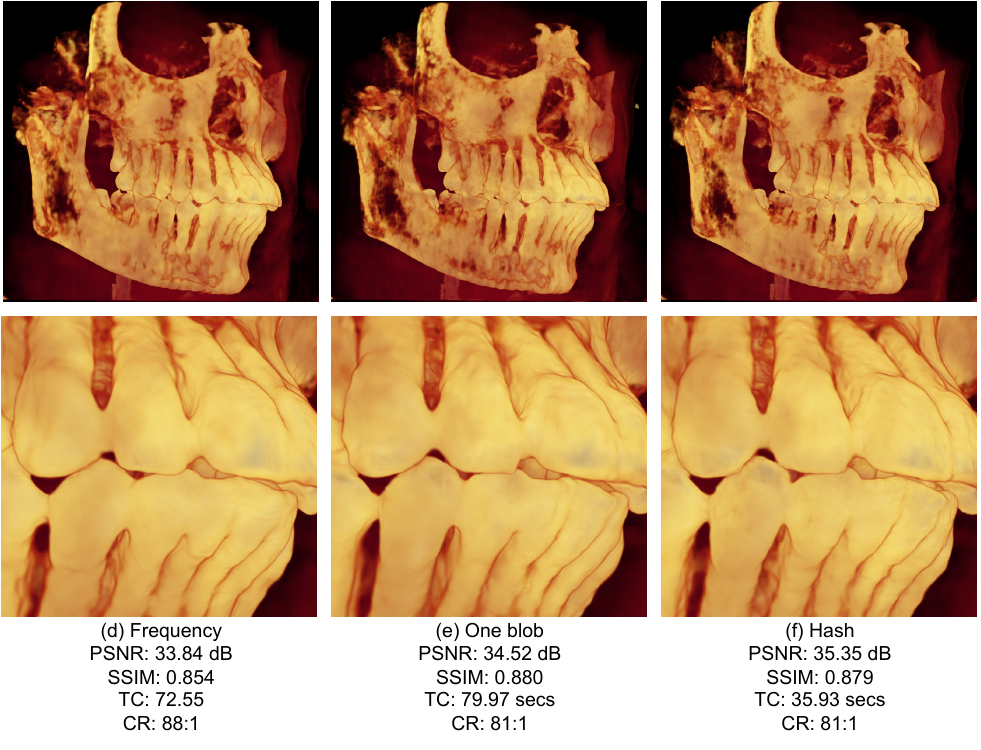}
    \end{subfigure}
    \caption{A comparison of various encoding schemes for compressing the skull dataset is presented. Each configuration was trained for 50 epochs, where each epoch is one complete pass over the entire volume. The top left image represents the ground truth rendering (a). The reconstruction quality is assessed using metrics such as PSNR and SSIM, while additional information including time to compress (TC) and compression ratio (CR) is provided beneath each image. Notably, the reconstructed rendering using Hash encoding demonstrates comparable quality to Triangle wave (c) and One blob encoding (e), but exhibits faster compression time, by at least 2x}
    \label{fig:OursVsEnc}
\end{figure*}

In the context of coordinate-based networks, the encoding of inputs holds significant importance, and we compare the performance of different encoding schemes to determine their effectiveness. These encoding schemes are essential for mapping the coordinate inputs to a higher-dimensional space. One such scheme is frequency encoding, where each coordinate $ x \in \mathbb{R} $ is represented as a sequence of sine and cosine functions : 

$E(x) = ( \sin(2^0\pi x), \cos(2^0\pi x) ... \sin(2^M\pi x), \cos(2^M\pi x) ),$

In our implementation, we select a value of $M=10$. Another encoding scheme, called triangle wave encoding, replaces the sine function with a more computationally efficient triangle wave and omits the cosine function. Additionally, we explore the one-blob encoding scheme, a generalized version of one-hot encoding, where a Gaussian kernel is applied over the normalized input coordinate and discretized into multiple bins. In our case, we use $k=64$ bins for this encoding scheme.

To showcase why we choose hash encoding over other schemes, we conduct a performance comparison of various encoding schemes for compressing the skull dataset. Figure \ref{fig:OursVsEnc} illustrates this comparison with similar compression ratios ranging from 81:1 to 91:1. We ensure that all networks have a similar number of trainable parameters. For the hash encoding network, we use a small MLP with 2 hidden layers and 64 neurons each, along with trainable multi-resolution hashtables $(L=6, W=8, T=2^{12})$. In contrast, the other networks consist of 12 hidden layers with 128 neurons each. We train the networks for 50 epochs where each epoch is one complete pass over the entire volume. When the network employed no encoding, specifically identity encoding, it struggled to preserve the high-frequency details present in the volume data, resulting in a smoothed approximation of the field data. Conversely, schemes such as frequency encoding, triangle wave encoding, and one-blob encoding enabled a more accurate representation of the field data within the same-sized network. However, these schemes still take substantial training times. In the case of multi-resolution hash encoding, the inclusion of trainable parameters within the hash table facilitated training with a smaller MLP architecture. Moreover, since the hash tables for all resolutions are queried in parallel, we achieved faster training times for compression. Notably, for our skull dataset, we observed a minimum 2x speedup with hash encoding compared to the alternative encoding schemes.



\begin{table}[htbp]
\centering
\caption{Datasets used in our evaluation.}
\label{tab:dataset}
\setlength{\tabcolsep}{2.5pt}
\begin{tabular}{c p{2.2cm} c c p{2.2cm}}
\hline
S.N. & Name         & Dimension     & Type & Source \\
\hline
1 & Tooth           & 103x94x161    & uint8 & Open scivis \cite{opensci}                   \\
2 & MRI ventricles  & 256x256x124   & uint8 & \cite{mriventricles}  \\
3 & MRHead          & 256x256x130   & uint16 & Slicer \cite{slicer}                         \\
4 & Aneurism        & 256x256x256   & uint8 & Philips Research  \cite{philips}             \\
5 & Skull           & 256x256x256   & uint8 & Siemens  \cite{skull}   \\
6 & Foot            & 256x256x256   & uint16 & Philips Research  \cite{philips}     \\
7 & Mrt-angio       & 416x512x112   & uint16 & Institute for Neuroradiology \cite{mrtangio}    \\
8 & Panoramix       & 441x321x215   & int16 & Slicer \cite{slicer}    \\
9 & CT-chest        & 512x512x139   & int32 & Slicer \cite{slicer}    \\
10 & CTA-cardio     & 512x512x321   & int16 & Slicer \cite{slicer}   \\
11 & Manix          & 512x512x460   & int16 & \cite{manix}   \\
12 & Boston teapot          & 256x256x178     & uint8 & Terarecon Inc \cite{boston}      \\
13 & Backpack               & 512*512*373     & uint16 & Viatronix Inc \cite{backpack}      \\
14 & Engine                 & 256x256x128     & uint8 & General Electric \cite{engine}   \\
15 & Tacc turbulence        & 256x256x256     & float32 & Open scivis \cite{opensci}   \\
16 & Csafe heptane          & 302x302x302     & uint16 & \cite{csafe}    \\
17 & Vorticity magnitude     & 480x720x120    & float32 & \cite{vorticity}  \\
18 & Magnetic reconnection  & 512x512x512     & float32 & \cite{magnetic_reconnection}    \\

\hline
\end{tabular}
\end{table}

\begin{figure*}[htbp]
  \centering

  \begin{minipage}{\textwidth}
    \begin{subfigure}[b]{0.19\textwidth}
      \centering
      \includegraphics[width=\textwidth]{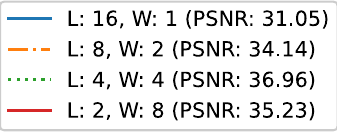}
    \end{subfigure}
    \hfill
    \begin{subfigure}[b]{0.19\textwidth}
      \centering
      \includegraphics[width=\textwidth]{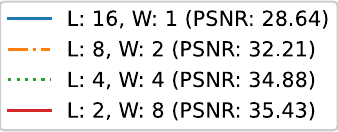}
    \end{subfigure}
    \hfill
    \begin{subfigure}[b]{0.19\textwidth}
      \centering
      \includegraphics[width=\textwidth]{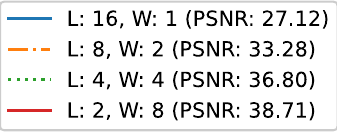}
    \end{subfigure}
    \hfill
    \begin{subfigure}[b]{0.19\textwidth}
      \centering
      \includegraphics[width=\textwidth]{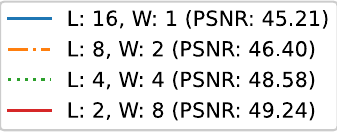}
    \end{subfigure}
    \hfill
    \begin{subfigure}[b]{0.19\textwidth}
      \centering
      \includegraphics[width=\textwidth]{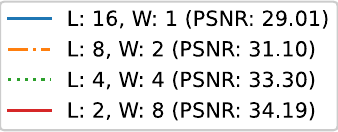}
    \end{subfigure}
    \hfill
  \end{minipage}

  \begin{minipage}{\textwidth}
    \begin{subfigure}[b]{0.19\textwidth}
      \centering
      \includegraphics[width=\textwidth]{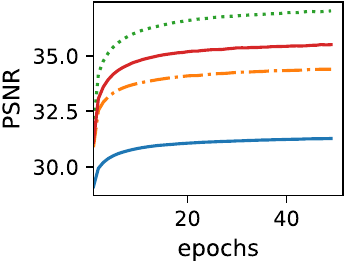}
      \caption{Csafe-heptane}
      \label{fig:hyperpara/csafe_heptane}
    \end{subfigure}
    \hfill
    \begin{subfigure}[b]{0.19\textwidth}
      \centering
      \includegraphics[width=\textwidth]{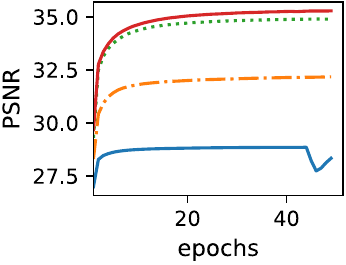}
      \caption{Panoramix-cropped}
      \label{fig:hyperpara/Panoramix-cropped}
    \end{subfigure}
    \hfill
    \begin{subfigure}[b]{0.19\textwidth}
      \centering
      \includegraphics[width=\textwidth]{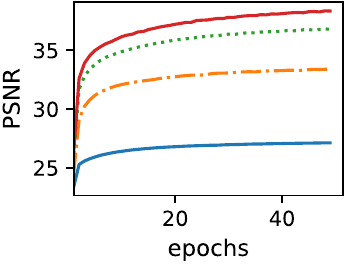}
      \caption{Engine}
      \label{fig:hyperpara/engine}
    \end{subfigure}
    \hfill    
    \begin{subfigure}[b]{0.19\textwidth}
      \centering
      \includegraphics[width=\textwidth]{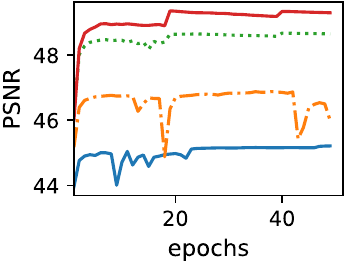}
      \caption{Tacc-turbulence}
      \label{fig:hyperpara/tacc_turbulence}
    \end{subfigure}
    \hfill 
    \begin{subfigure}[b]{0.19\textwidth}
      \centering
      \includegraphics[width=\textwidth]{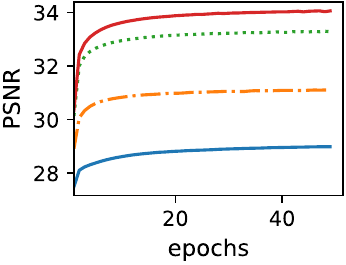}
      \caption{Skull}
      \label{fig:hyperpara/skull}
    \end{subfigure}
    \hfill 
  \end{minipage}

  \caption{Figure shows compression error against the number of epochs for various combinations of $L$ and $W$ while keeping the total encoding parameters as constant. Notably, we find that configurations with $W$ values ranging from 4 to 8 consistently achieve higher PSNR values across all datasets.}
  \label{fig:hyperpara-LvsW}
\end{figure*}

\subsection{Hyperparameter study}

Multi-resolution hash encoding offers flexibility in determining the number of encoding parameters, which is given by the product of hyperparameters $L$ (number of levels), $W$ (number of weights in each entry of the hash table), and $T$ (size of the hash table). The choice of the hash table size $T$ involves a trade-off between compression performance, memory usage, and compression quality. For $T$, we rely on the findings reported by M{\"{u}}ller \emph{et al.} \cite{instantnerf}. Based on their results, higher values of T lead to improved reconstruction quality, but at the expense of increased memory usage and slower training and inference. The memory footprint is linear in $T$, whereas quality and performance tend to scale sub-linearly. In our experiments, we select T values ranging from $2^8$ to $2^{12}$ for different volume datasets, to lower the training and inference time while still having acceptable compression quality.

The hyperparameters $L$ and $W$ also influence the trade-off between compression quality and performance. To determine the optimal range of $L$ and $W$ for our specific use case, we plot the compression error against the number of epochs for various combinations of $L$ and $W$, while keeping the total number of encoding parameters fixed (i.e., maintaining a constant compression ratio). As depicted in Figure \ref{fig:hyperpara-LvsW}, a configuration with $W$ between 4 and 8 appears to yield favorable results across most datasets, and thus, we adopt this configuration for the majority of our evaluations. We vary $L$ to achieve different compression levels while keeping the total encoding parameters constant.

\begin{figure*}[htbp]
  \centering
    \begin{subfigure}{0.2\textwidth}
      \centering
      \includegraphics[width=\textwidth]{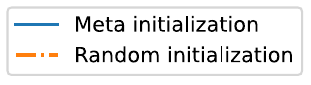}
      \label{fig:legend-medical}
    \end{subfigure}
    \hfill
  
  \begin{minipage}{\textwidth}
    \begin{subfigure}[b]{0.19\textwidth}
      \centering
      \includegraphics[width=\textwidth]{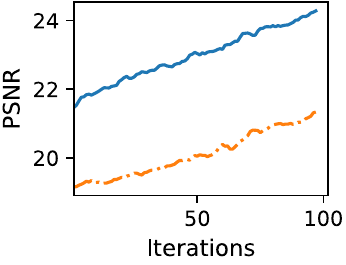}
      \label{fig:Mri_ventricles-100}
    \end{subfigure}
    \hfill
    \begin{subfigure}[b]{0.19\textwidth}
      \centering
      \includegraphics[width=\textwidth]{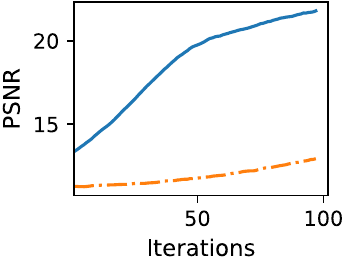}
      \label{fig:Panoramix-cropped-100}
    \end{subfigure}
    \hfill
    \begin{subfigure}[b]{0.19\textwidth}
      \centering
      \includegraphics[width=\textwidth]{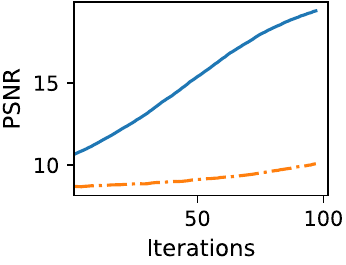}
      \label{fig:CT-chest-100}
    \end{subfigure}
    \hfill    
    \begin{subfigure}[b]{0.19\textwidth}
      \centering
      \includegraphics[width=\textwidth]{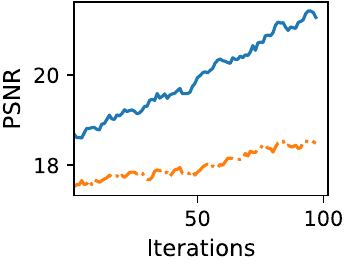}
      \label{fig:Foot-100}
    \end{subfigure}
    \hfill 
    \begin{subfigure}[b]{0.19\textwidth}
      \centering
      \includegraphics[width=\textwidth]{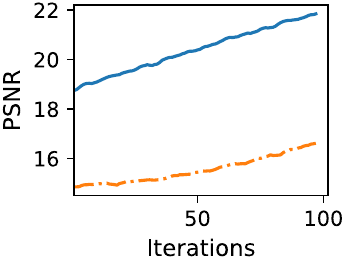}
      \label{fig:MRHead-100}
    \end{subfigure}
    \hfill 
  \end{minipage}

  \begin{minipage}{\textwidth}
    \begin{subfigure}[b]{0.19\textwidth}
      \centering
      \includegraphics[width=\textwidth]{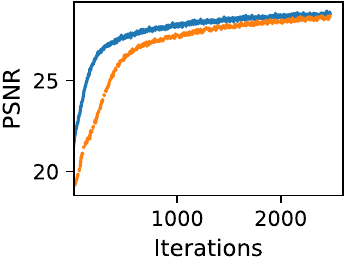}
      \caption{MRI ventricles}
      \label{fig:Mri_ventricles-2500}
    \end{subfigure}
    \hfill
    \begin{subfigure}[b]{0.19\textwidth}
      \centering
      \includegraphics[width=\textwidth]{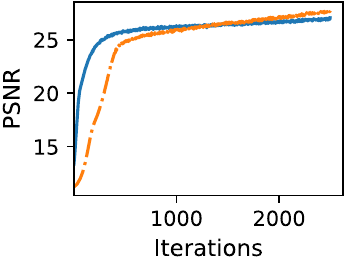}
      \caption{Panoramix cropped}
      \label{fig:Panoramix-cropped-2500}
    \end{subfigure}
    \hfill
    \begin{subfigure}[b]{0.19\textwidth}
      \centering
      \includegraphics[width=\textwidth]{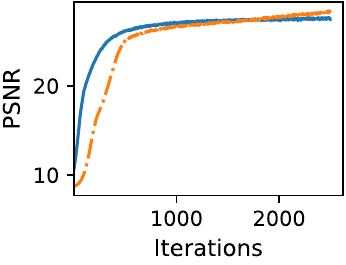}
      \caption{CT-chest}
      \label{CT-chest-2500}
    \end{subfigure}
    \hfill
    \begin{subfigure}[b]{0.19\textwidth}
      \centering
      \includegraphics[width=\textwidth]{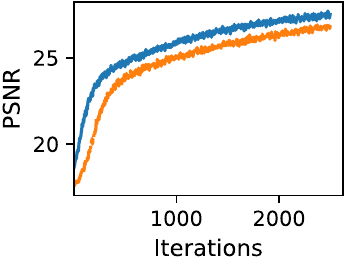}
      \caption{Foot}
      \label{fig:Foot-2500}
    \end{subfigure}
    \hfill
    \begin{subfigure}[b]{0.19\textwidth}
      \centering
      \includegraphics[width=\textwidth]{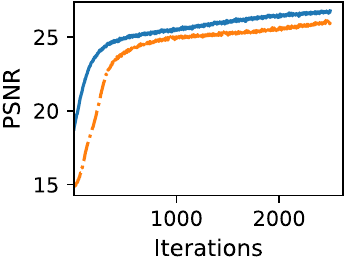}
      \caption{MRHead}
      \label{fig:MRHead-2500}
    \end{subfigure}
    \hfill
  \end{minipage}
  
  \caption{Comparison of convergence speed between meta-learned initialization and random initialization for intra-domain weight transfer. The reconstruction PSNR is reported for the first 100 iterations (top row) and 2500 iterations (bottom row) for each dataset. The number of iterations corresponds to the number of gradient updates performed during the training process. The meta-learned approach exhibits faster convergence, particularly evident at the initial training phase. While the random initialization approach eventually achieves a similar PSNR to the meta-learned approach, the latter surpasses random initialization after only 100 iterations in terms of PSNR.}
  \label{fig:meta-medical}
\end{figure*}

\begin{figure*}[htbp]
  \centering
    \begin{subfigure}{0.2\textwidth}
      \centering
      \includegraphics[width=\textwidth]{figures/plots/meta/legend.pdf}
      \label{fig:legend-non-medical}
    \end{subfigure}
    \hfill
  
  \begin{minipage}{\textwidth}
    \begin{subfigure}[b]{0.19\textwidth}
      \centering
      \includegraphics[width=\textwidth]{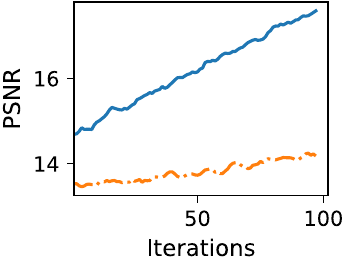}
      \label{fig:Engine-100}
    \end{subfigure}
    \hfill
    \begin{subfigure}[b]{0.19\textwidth}
      \centering
      \includegraphics[width=\textwidth]{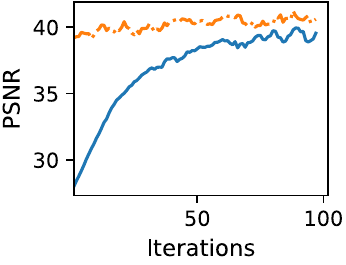}
      \label{fig:Tacc_turbulence-100}
    \end{subfigure}
    \hfill
    \begin{subfigure}[b]{0.19\textwidth}
      \centering
      \includegraphics[width=\textwidth]{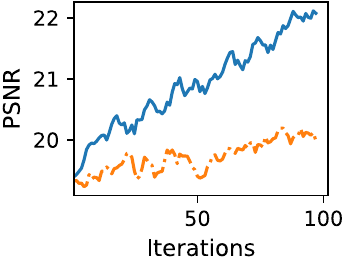}
      \label{fig:Csafe_heptane-100}
    \end{subfigure}
    \hfill    
    \begin{subfigure}[b]{0.19\textwidth}
      \centering
      \includegraphics[width=\textwidth]{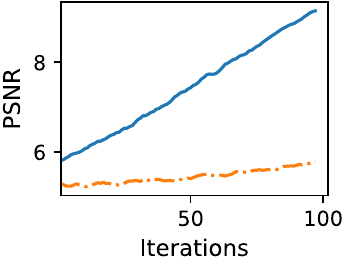}
      \label{fig:Velocity-mag-100}
    \end{subfigure}
    \hfill 
    \begin{subfigure}[b]{0.19\textwidth}
      \centering
      \includegraphics[width=\textwidth]{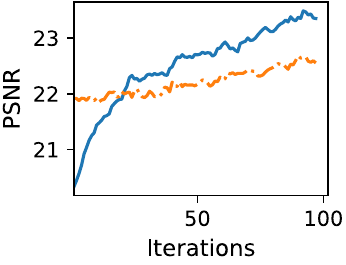}
      \label{fig:Boston_teapot-100}
    \end{subfigure}
    \hfill 
  \end{minipage}

  \begin{minipage}{\textwidth}
    \begin{subfigure}[b]{0.19\textwidth}
      \centering
      \includegraphics[width=\textwidth]{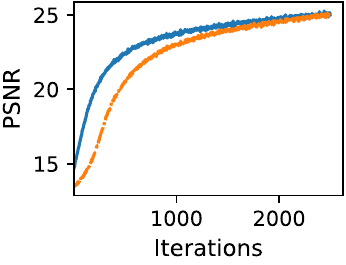}
      \caption{Engine}
      \label{fig:Engine}
    \end{subfigure}
    \hfill
    \begin{subfigure}[b]{0.19\textwidth}
      \centering
      \includegraphics[width=\textwidth]{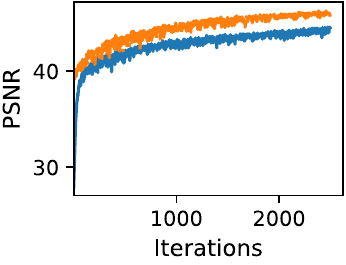}
      \caption{Tacc turbulence}
      \label{fig:Tacc_turbulence}
    \end{subfigure}
    \hfill
    \begin{subfigure}[b]{0.19\textwidth}
      \centering
      \includegraphics[width=\textwidth]{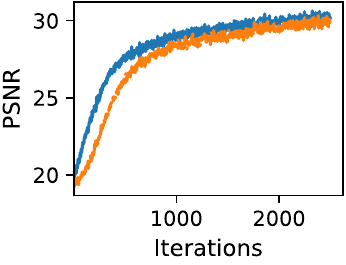}
      \caption{Csafe heptane}
      \label{fig:Csafe_heptane}
    \end{subfigure}
    \hfill
    \begin{subfigure}[b]{0.19\textwidth}
      \centering
      \includegraphics[width=\textwidth]{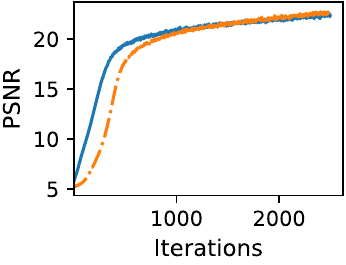}
      \caption{Vorticity mag}
      \label{fig:Velocity-mag}
    \end{subfigure}
    \hfill
    \begin{subfigure}[b]{0.19\textwidth}
      \centering
      \includegraphics[width=\textwidth]{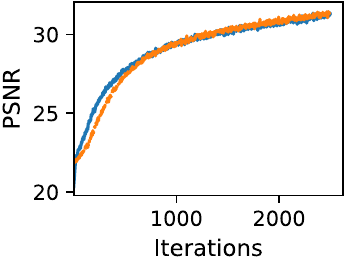}
      \caption{Boston teapot}
      \label{fig:Boston_teapot}
    \end{subfigure}
    \hfill
  \end{minipage}
  
  \caption{Comparison of convergence speed between meta-learned initialization and random initialization for inter-domain weight transfer. The reconstruction PSNR is reported for the first 100 iterations (top row) and 2500 iterations (bottom row) for each dataset. The number of iterations corresponds to the number of gradient updates performed during the training process. The meta-learned approach demonstrates slightly faster convergence for datasets a) engine, c) Csafe heptane, and d) Vorticity magnitude. However, it provides minimal to no advantage for dataset e) Boston teapot and performs poorly for dataset b) Tacc turbulence.}
  \label{fig:meta-non-medical}
\end{figure*}

\subsection{Performance Comparison with Meta-Initialization}

In this section, we investigate the potential benefits of incorporating meta-learned initialization into our volume representation network. Initially, we evaluate the impact of meta-initialization within a specific distribution of volume data. We perform meta-learning with a particular set of medical volume data and then use the learned parameters to compress another medical volume data that was not included in the meta-learning stage. Additionally, we also investigate whether meta-learned initialization from one domain, such as medical volume data, can enhance the performance of the network when applied to compress volume data from a different domain, such as scientific simulations.

\subsubsection{Intra-domain Weight Transfer}
For the first experiment, we apply meta-learning to optimize coordinate-based networks for representing medical datasets. The underlying dataset for meta-learning consists of medical volume data \# 1 to 10 from Table \ref{tab:dataset}. We use Reptile learning to meta-learn the initial weights for each medical volume data shown in Figure \ref{fig:meta-medical}. For each experiment, we hold out one of the volume data as testing data and perform meta-learning using the remaining datasets. For every iteration of the meta-learning stage, we randomly sample a volume data from the dataset pool and perform k gradient updates on that sample before updating the initial parameters using the update rule outlined in Algorithm \ref{alg:reptile}. Here, k gradient updates correspond to completing one full pass over the entire volume. 

At testing time, we optimize a similar-sized network initialized with meta-learned parameters to compress the test dataset. For comparison, we also optimize another network with random initialization for the same test dataset. The underlying MLP architecture consists of 2 hidden layers with 64 neurons each, and we apply multi-resolution hash encoding with $L=6, W=4,$ and $T=2^{12}$. As seen in Figure \ref{fig:meta-medical}, using the learned initial weights enables faster convergence, which is particularly evident at the initial training phase. While the random initialization approach eventually achieves a similar PSNR to the meta-learned approach, the latter surpasses random initialization after only a few gradient updates.

\subsubsection{Inter-domain Weight Transfer}
In the second experiment, we utilize the same set of medical volume datasets (datasets \# 1 to 10 from Table \ref{tab:dataset}) to generate meta-learned parameters. Subsequently, during the testing phase, we use these meta-learned parameters to optimize a coordinate-based network for compressing different volume datasets that are not from the medical domain. The comparison with random initialization for these datasets is presented in Figure \ref{fig:meta-non-medical}. We observe that, with the exception of Tacc turbulence and Boston teapot datasets where the benefits of meta-learned initialization were slight or insignificant, our method typically resulted in faster convergence across the majority of the datasets. These two evaluations suggest that leveraging domain-specific knowledge through meta-learned initialization can bring significant improvements to the efficiency of volumetric data compression and representation.

\begin{figure*}[htbp]
  \centering
  \includegraphics[width=\linewidth]{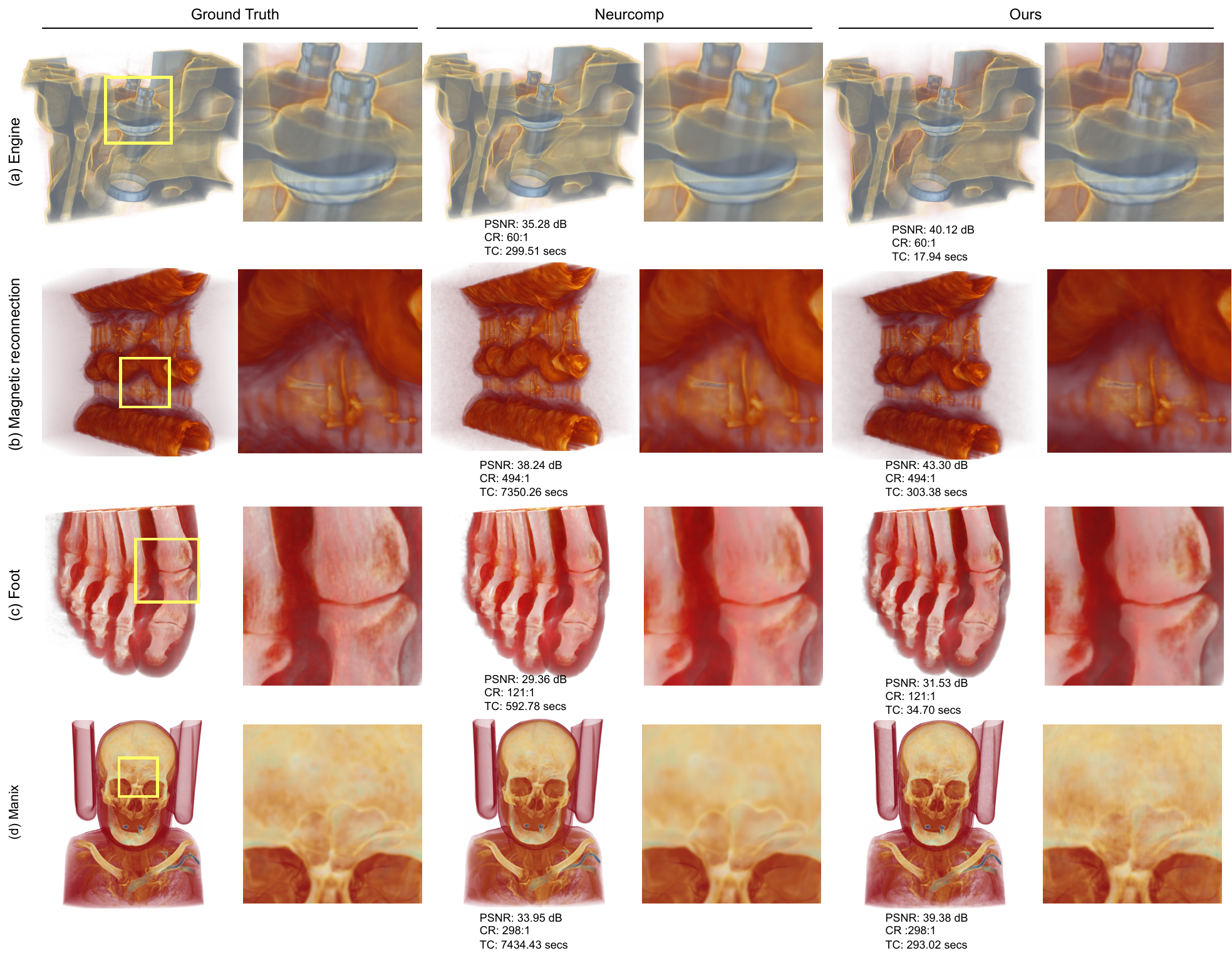}
  \centering
  \caption{Comparative Compression Analysis - Our Approach vs Neurcomp. The first column presents the ground truth renderings of four distinct datasets: a) Engine, b) Magnetic Reconnection, c) Foot, and d) Manix. The second column showcases the renderings using Neurcomp, while the third column highlights the results achieved with our proposed method. This visual comparison shows the effectiveness of our approach in compressing diverse datasets.}   
  \label{fig:OursVsNeurcomp}

\end{figure*}

\begin{figure}[htbp]
  \centering
  \includegraphics[width=\linewidth]{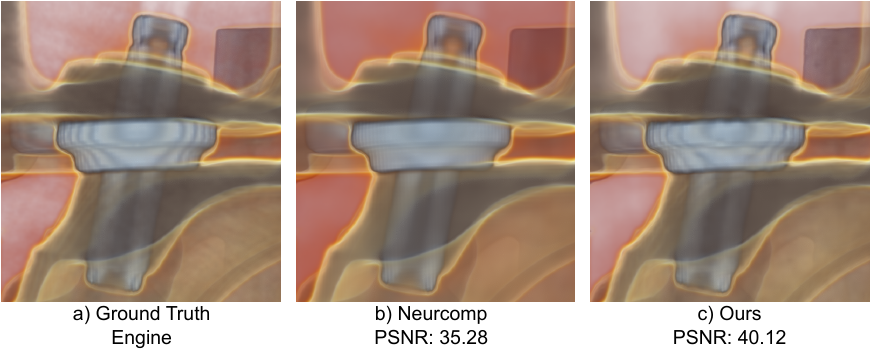}
  \centering
  \caption{Visual comparison between Neurcomp and Ours for the engine dataset.}   
  \label{fig:OursVsNeurcomp-2}

\end{figure}

\begin{table*}[htbp]
\centering
\caption{Comparison of compression results for different datasets using Neurcomp, Tthresh and our method. (CR: Compression ratio, TC: Time to compress in seconds, GT: Ground Truth, FPS: Frames per second). Here GT FPS means the frames per second we observe from rendering the ground truth volume data. } 
\label{tab:tthresh}

\begin{tabular}{|c|c|c|c|c|c|c|c|c|c|c|c|}
\hline
\textbf{Dataset} & \multicolumn{3}{c|}{\textbf{Neurcomp}} & \multicolumn{3}{c|}{\textbf{Tthresh}} & \multicolumn{4}{c|}{\textbf{Ours}} & \multicolumn{1}{c|}{\textbf{GT}}\\
\cline{2-11}
 & \textbf{PSNR} & \textbf{CR} & \textbf{TC} & \textbf{PSNR} & \textbf{CR} & \textbf{TC} & \textbf{PSNR} & \textbf{CR} & \textbf{TC} & \textbf{FPS} & \textbf{FPS}\\
\hline

Tooth                  & 34.52 & 40:1  & 36.58   & 34.13 & 133:1 & 4.04   & 33.92 & 40:1  & 3.89    & 24 & 53 \\
MRI ventricles         & 22.19 & 58:1  & 289.05  & 25.10 & 309:1 & 1.48   & 24.69 & 58:1  & 19.84   & 23 & 47 \\
MRHead                 & 22.78 & 61:1  & 302.72  & 26.21 & 230:1 & 1.53   & 25.75 & 61:1  & 21.02   & 19 & 46 \\
Aneurism               & 31.64 & 121:1 & 596.20  & 38.83 & 35:1  & 3.43   & 39.46 & 121:1 & 40.43   & 21 & 52 \\
Skull                  & 30.43 & 121:1 & 593.18  & 34.41 & 80:1  & 3.33   & 34.88 & 121:1 & 40.27   & 19 & 40 \\
Foot                   & 29.36 & 121:1 & 592.78  & 31.43 & 56:1  & 3.30   & 31.53 & 121:1 & 34.70   & 17 & 50 \\
Mrt-angio              & 27.39 & 172:1 & 850.60  & 31.16 & 64:1  & 5.90   & 31.80 & 172:1 & 48.04   & 17 & 51 \\
Panoramix              & 27.35 & 220:1 & 1095.98 & 35.60 & 162:1 & 8.60   & 35.07 & 220:1 & 107.63  & 22 & 57 \\
CT-chest               & 40.78 & 67:1  & 4849.20 & 40.10 & 75:1  & 8.52   & 39.95 & 67:1  & 95.02   & 13 & 42 \\
CTA-cardio             & 35.99 & 208:1 & 5220.44 & 38.61 & 221:1 & 19.61  & 38.36 & 208:1 & 205.45  & 16 & 48 \\
Manix                  & 33.95 & 298:1 & 7434.43 & 39.69 & 341:1 & 20.446 & 39.38 & 298:1 & 293.02  & 18 & 53 \\
Boston teapot          & 35.16 & 42:1  & 622.73  & 40.19 & 396:1 & 1.95   & 39.68 & 42:1  & 25.74   & 18 & 44 \\
Backpack               & 36.62 & 55:1  & 12918.76& 37.15 & 55:1  & 16.13  & 38.77 & 55:1  & 394.41  & 17 & 47 \\
Engine                 & 35.28 & 60:1  & 299.51  & 40.70 & 128:1 & 1.34   & 40.12 & 60:1  & 17.94   & 18 & 51\\
Tacc turbulence        & 37.72 & 528:1 & 421.91  & 44.37 & 464:1 & 2.59   & 45.73 & 528:1 & 10.93   & 19 & 49 \\
Csafe-heptane          & 37.11 & 134:1 & 1199.07 & 40.78 & 203:1 & 5.5    & 40.83 & 134:1 & 60.77   & 18 & 55 \\
Vorticity magnitude    & 29.10 & 152:1 & 2079.05 & 29.64 & 85:1  & 38.87  & 30.14 & 152:1 & 92.14   & 16 & 52 \\
Magnetic reconnection  & 38.24 & 494:1 & 7350.26 & 43.25 & 470:1 & 20.91  & 43.3  & 494:1 & 303.38  & 31 & 55 \\

\hline
\end{tabular}
\end{table*}

\subsection{Comparison with State of the Art Volume Compression Methods}

In this section, we primarily benchmark our approach against Neurcomp, which represents a state-of-the-art in neural compression technique. Additionally, we perform comparisons with Tthresh, an advanced CPU-based volume compression technique. To assess the effectiveness of our volumetric neural representation in encoding the ground truth volume data, we sequentially decode the volume at its original resolution from the compressed representation, and measure the similarity between the original volume data and the volume predicted by our methodology, Neurcomp, and Tthresh. 

For the comparisons we make with the two baselines: Neurcomp and Tthresh, we opt to not use meta-learning. This decision is based on our observations from sections 4.3.1 and 4.3.2, which indicated that the benefits of meta-learned initialization might be minimal or insignificant for certain scenarios, particularly for inter-domain weight transfer. Since the datasets we use for evaluation in this section come from different domains, we choose to make the comparisons without using meta-learning. Thus, the metric "TC" (time to compress) solely represents the time required for the actual compression process, excluding the meta-learning stage.

Furthermore, we use different transfer functions for different datasets because the optimal transfer function for a particular dataset may vary depending on the characteristics of the data. However, to ensure a fair comparison, we use the same transfer function for comparing our method with ground truth, neurcomp, and tthresh. We also represent the ground truth data in single precision floating point format for all comparisons. This guarantees that our approach accurately reflects the performance compared to the baseline techniques.

\subsubsection{Comparison with Neurcomp}

In order to provide a comprehensive comparison between our method and Neurcomp \cite{neurcomp}, we execute our method on different datasets for varying compression ratios, as shown in Figure \ref{fig:OursVsNeurcomp} and table \ref{tab:tthresh}. Both methods were run under the same conditions, utilizing a batch size of $2^{14}$ and the networks were trained for a total of 50 epochs, with each epoch representing a complete pass over the entire volume of data.The underlying architecture we employed for the comparison encompasses two hidden layers, each containing 64 neurons. For input encoding, we use multi-resolution hash tables with parameters $W=8$, and $T=2^{12}$, and we vary $L$ between $4$ and $12$ to reflect different compression ratios. To ensure equivalent compression ratios across all datasets for both methods, we use an 8-layer network for Neurcomp, while adjusting the number of neurons accordingly. This adjustment ensured that both networks achieved the same level of compression.

A qualitative analysis reveals that the rendered images produced by both Neurcomp and our method look similar for the same compression ratios. However, our approach outperforms Neurcomp in terms of PSNR, indicating superior image quality. Upon closer examination, we observe that Neurcomp has a tendency to generate smoother surfaces in the resulting renderings, which may at times lead to an under-representation of the high-frequency noise inherent in the original, ground truth data. This phenomenon is particularly noticeable in the case of the engine dataset, as depicted in Figure \ref{fig:OursVsNeurcomp-2}. Our method, conversely, exhibits better capability to preserve and represent this high-frequency noise, leading to a more accurate representation of the ground truth data,

\subsubsection{Comparison with Tthresh}

In order to assess our method in comparison to Tthresh, we modify the input parameters of Tthresh. As it accepts an error parameter (for instance, PSNR) rather than a specific compression ratio, we adjust the PSNR values for Tthresh roughly equivalent to those obtained from our approach and compare the compression ratios. 

The quantitative outcomes of our approach versus Tthresh are presented in Table \ref{tab:tthresh}. Our analysis reveals that the performance of the two methods is data-dependent: for certain datasets, our method outperforms Tthresh, while for others, Tthresh achieves superior results. While the PSNR values obtained with Tthresh are comparable to those achieved by our method, a key distinction exists in the post-compression handling of data. Unlike Tthresh, our method does not require data decompression prior to rendering; instead, we can directly perform ray-traced direct volume rendering from the compressed representation. The observed frames per second (FPS) for rendering different volume data with our method is shown in Table \ref{tab:tthresh}. Our approach treats the neural representation of data as a compressed version of the original dataset, which significantly reduces memory usage. Moreover, this method permits flexible sampling without the necessity for explicit interpolation. 

\begin{figure}[htbp]
  \centering
  \includegraphics[width=\linewidth]{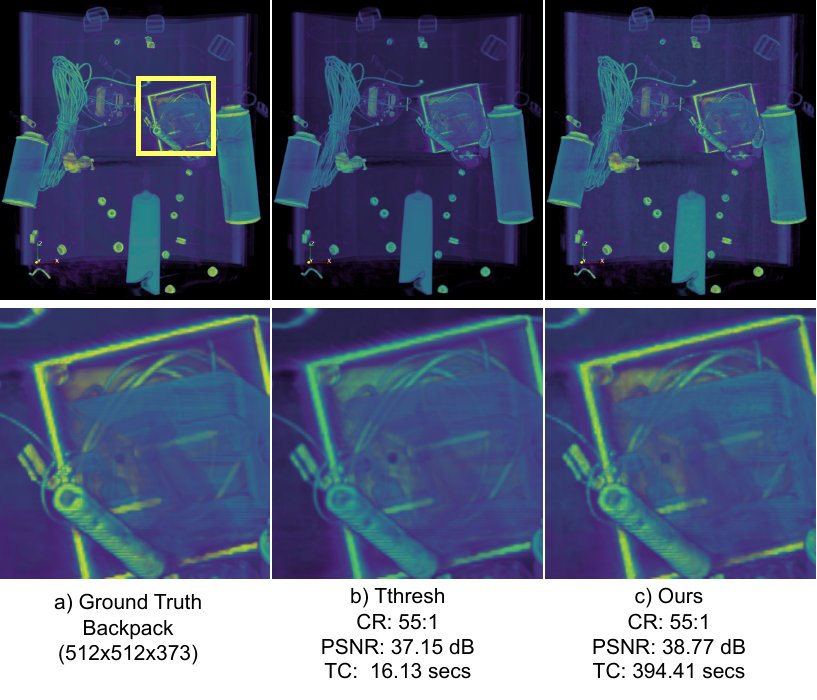}
  \centering
  \caption{Comparison with Thresh for lower compression ratio for Backpack dataset}   
  \label{fig:Tthresh-backpack}

\end{figure}

\begin{figure}[htbp]
  \centering
  \includegraphics[width=\linewidth]{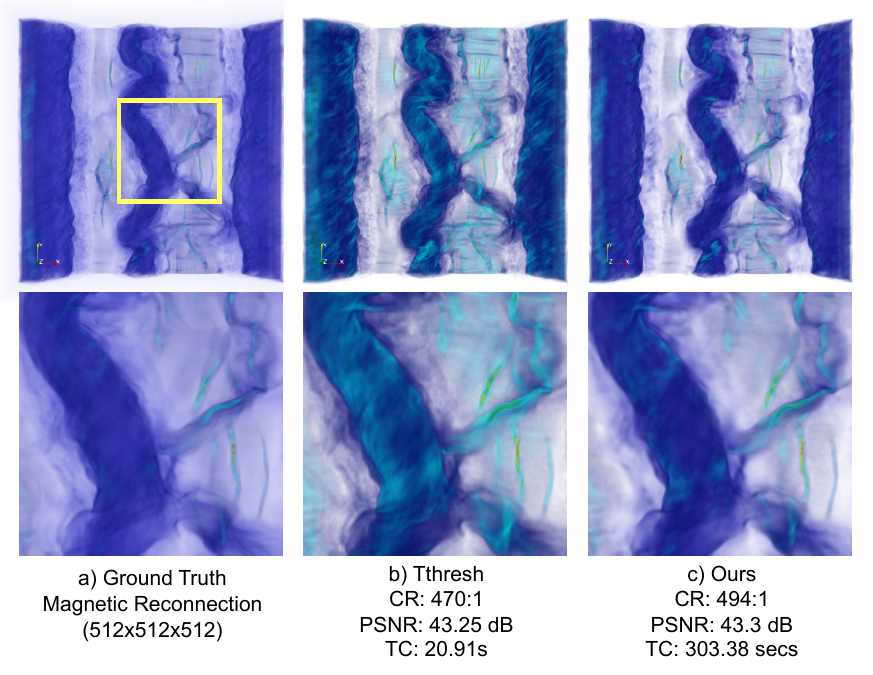}
  \centering
  \caption{Comparison with Thresh for higher compression ratio for Magnetic reconnection dataset}   
  \label{fig:Tthresh-mag}

\end{figure}

Figure \ref{fig:Tthresh-backpack} and Figure \ref{fig:Tthresh-mag}, present a comparison between Tthresh and our method for the backpack and magnetic reconnection datasets. In these examples, our approach demonstrates a slight advantage over Tthresh, achieving similar PSNR values but with better compression ratios.

\begin{figure}[htbp]
  \centering
  \includegraphics[width=0.9\linewidth]{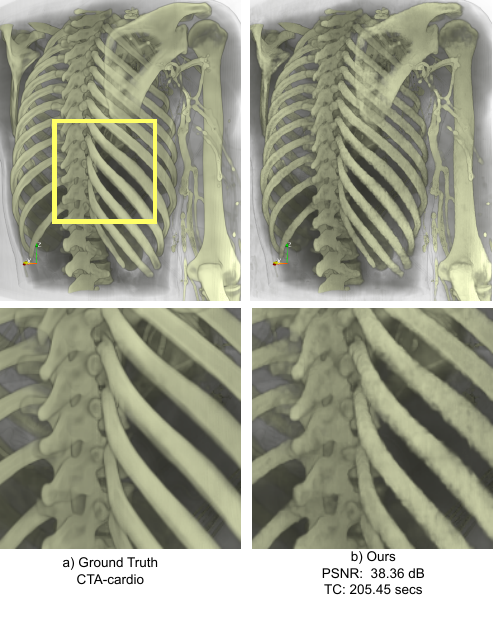}
  \centering
  \caption{Ground truth compared to ours for CTA cardio dataset. Unwanted artifacts can be seen on the bone structure. Compression ratio: 208:1 }   
  \label{fig:cta-cardio-fail}

\end{figure}

\begin{table}[htbp]
\centering
\caption{Results for CTA-cardio dataset: with and without gradient loss. }
\label{tab:cta-cardio-fail}
\begin{tabular}{|c|c|c|}
\hline
 & \textbf{Without grad. loss} & \textbf{With grad. loss} \\
\hline
\textbf{PSNR}        & 38.36 dB   & 37.45 dB \\
\hline
\textbf{Grad. loss}  & 0.3517     & 0.3728 \\
\hline
\end{tabular}
\end{table}

\subsection{Limitations}
While our approach shows promise, there are some limitations to consider. For some datasets, particularly at higher compression ratios, we observed the introduction of noise and artifacts in the compressed volume. An example of this can be seen in Figure \ref{fig:cta-cardio-fail} for the CTA cardio dataset at a high compression ratio of 208:1, where visually undesirable surface roughness is present on the bone structure. 

To address this issue, we explored the incorporation of a small percentage of gradient mean squared error (MSE) loss in our total loss function, as gradient-based volume rendering techniques play a crucial role in accurate visualization of complex structures. In particular, we implement the central difference method to compute the gradients for the predicted volume and the ground truth volume. While we were able to notice an improvement in the overall gradient MSE loss, it came at the expense of lower PSNR, as shown in Table \ref{tab:cta-cardio-fail}. This also aligns with the findings reported by Lu \emph{et al.}\cite{neurcomp}. Further exploration is needed to fully understand and mitigate these trade-offs.

\section{Discussion and Conclusion}

In this work, we proposed a volume representation network based on coordinate-based networks and multi-resolution hash encoding. Our method achieved efficient and high-quality compression of volumetric data by leveraging spatial encoding and trainable hash tables. Our comparison to the existing neural compression method showcased our method's superior performance in terms of both quality of representation and time to convergence.

We also introduced the concept of meta-learned initialization for volume compression, which leverages domain-specific knowledge to enhance the efficiency of neural volume representation. While we found that the benefits of meta-learned initialization are highly data-dependent, we observed consistent improvements across a broad range of datasets, suggesting that our approach can bring general benefits in a variety of contexts. 

Looking ahead, there are several potential directions for future work. While our current method treats the entire volume as a single entity, future work could look into developing more sophisticated models that are capable of recognizing and separately handling different regions within the volume data. This would allow for region-specific compression that could potentially yield even better compression ratios and visual quality. We hope our work will pave the way for further research into the use of meta-learning for volume data compression and opens new opportunities for efficient volume data management.

Another interesting avenue for future work is the exploration of hybrid compression techniques that combine the strengths of neural network-based compression with traditional compression algorithms. Integrating neural network-based methods with existing compression techniques, such as lossless or lossy compression algorithms, could potentially yield even better compression ratios and preservation of fine details.

\bibliographystyle{eg-alpha-doi}  
\bibliography{NeuralVolumeRepresentation}        




\end{document}